\documentclass[a4paper,fleqn]{cas-sc}

\usepackage[numbers]{natbib}
\usepackage{pifont}
\usepackage{placeins}

\def\tsc#1{\csdef{#1}{\textsc{\lowercase{#1}}\xspace}}
\tsc{WGM}
\tsc{QE}
\tsc{EP}
\tsc{PMS}
\tsc{BEC}
\tsc{DE}

\begin{document}
\let\WriteBookmarks\relax
\def\floatpagepagefraction{1}
\def\textpagefraction{.001}
\shorttitle{Interpretable CPTabKAN for EEG-Based MCI Detection}
\shortauthors{Yosef B. Wirian and Qiang Cheng}

\title [mode = title]{Interpretable Concept-Guided Polynomial Tabular Kolmogorov-Arnold Network for EEG-Based Mild Cognitive Impairment Detection}

\author[1]{Yosef B. Wirian}[type=editor,
                        auid=000,bioid=1,
                        orcid=0009-0006-0051-891X]

\ead{yosef.wirian@uky.edu}

\credit{Conceptualization, Data curation, Formal analysis, Investigation, Methodology, Project administration, Software, Validation, Visualization, Writing – original draft, Writing – review \& editing}

\affiliation[1]{organization={Computer Science Department, University of Kentucky},
                addressline={351 Ralph G. Anderson Building}, 
                city={Lexington},
                postcode={40506}, 
                state={KY},
                country={USA}}

\author[1,2]{Qiang Cheng}[type=editor,
                        auid=000,bioid=2,
                        orcid=0000-0002-3596-2838]

\cormark[1]

\ead{qiang.cheng@uky.edu}

\credit{Conceptualization, Formal analysis, Funding acquisition, Project administration, Resources, Supervision, Validation, Writing – original draft, Writing – review \& editing}

\affiliation[2]{organization={Institute for Biomedical Informatics, University of Kentucky},
                addressline={760 Press Ave}, 
                postcode={40508}, 
                city={Lexington},
                state={KY},
                country={USA}}

\cortext[cor1]{Corresponding author}

\makeatletter
\def\ps@first{%
  \let\@oddhead\@empty
  \let\@evenhead\@empty
  \def\@oddfoot{\reset@font\hfil\thepage\hfil}%
  \let\@evenfoot\@oddfoot
}
\def\ps@pprintTitle{%
  \let\@oddhead\@empty
  \let\@evenhead\@empty
  \def\@oddfoot{\reset@font\hfil\thepage\hfil}%
  \let\@evenfoot\@oddfoot
}
\makeatother
\pagestyle{plain}

\begin{abstract}
Early and scalable detection of mild cognitive impairment (MCI) remains an unresolved clinical challenge. Existing EEG-based screening approaches are constrained by handcrafted feature pipelines that discard neurophysiologically meaningful domain structure and deep learning classifiers that sacrifice interpretability for performance. No existing work unifies physiologically organized concept encoders, cross-concept interaction modeling, and nonlinear tabular classification in a sleep EEG-based MCI detection framework. This study proposes Concept-guided Polynomial-transformed Tabular learning using Kolmogorov-Arnold Network (CPTabKAN), which maps heterogeneous EEG-derived features into domain-informed concept representations, expands them via degree-2 polynomial transformation to expose first- and second-order interactions, and applies a Fourier-parameterized TabKAN classifier to learn nonlinear decision boundaries. CPTabKAN was evaluated on the Study of Osteoporotic Fractures cohort (372 subjects, overnight polysomnography), using 1,379 features organized into ten physiologically motivated concept groups. Under 10-fold cross-validation, CPTabKAN-Second Order achieved a weighted F1-score of 0.9038 (SD 0.034), outperforming GradientBoosting by 5.65 percentage points ($t(9)=1.934,p=0.043$, one-sided paired test), with advantages persisting under SMOTE-based balancing. Ablation analysis confirmed independent contributions from each component. Concept importance analysis revealed that power spectral density, multi-scale entropy, and Hjorth parameters dominated first-order weights, while cross-concept interactions involving Lempel-Ziv-Welch complexity, statistics, demographics, and slow oscillations exceeded all first-order scores. These results demonstrate that concept-structured, interaction-aware tabular learning surfaces physiologically coherent reasoning, supporting clinical trust.
\end{abstract}



\begin{keywords}
Mild cognitive impairment \sep Sleep EEG \sep Concept bottleneck learning \sep Kolmogorov-Arnold networks \sep Tabular biomedical classification
\end{keywords}

\maketitle

\section{Introduction}
Mild cognitive impairment (MCI) occupies a clinically critical intermediate stage between normal cognitive aging and dementia, during which early identification may support timely monitoring and intervention planning. Despite its importance, scalable and accessible early screening remains elusive. Neuroimaging modalities and cerebrospinal-fluid biomarkers offer diagnostic sensitivity but entail prohibitive costs, demand specialized infrastructure, and are impractical for population-level deployment. Neuropsychological assessments provide useful cognitive snapshots but lack sensitivity to the underlying neural dynamics that precede overt cognitive decline. Sleep electroencephalography (EEG) addresses several of these constraints simultaneously: it is non-invasive, relatively low-cost, portable across clinical settings from specialized laboratories to community health centers, and captures neural activity during naturally occurring brain states that may reveal subtle pathological changes not apparent during waking assessments. Sleep is closely linked to memory consolidation, neural synchrony, and neurodegenerative processes, and MCI-related alterations manifest in spectral power, signal complexity, slow oscillations, sleep spindles, and their interactions, making overnight sleep EEG a particularly well-motivated modality for cognitive impairment screening.

A substantial body of work has established the feasibility of EEG-based MCI classification across a range of methodological families. Traditional classifiers, including support vector machines (SVM), $k$-nearest neighbors (KNN), and logistic regression, remain competitive on small clinical datasets due to their relative stability, but their performance is bounded by the quality of feature engineering and their limited capacity to model nonlinear interactions among heterogeneous EEG descriptors~\cite{geng2022sleep,Aljalal2024,diagnostics14151619}. Tree-based ensemble methods such as random forests~\cite{rutkowski2024mild}, XGBoost, and LightGBM often perform strongly on tabular biomedical data by capturing nonlinear effects through sequential boosting. Deep learning architectures, including bidirectional LSTM, bidirectional GRU, and hybrid convolutional-recurrent models such as DCNN-SBiL, can learn richer temporal and hierarchical representations~\cite{geng2022sleep,Said2024,devi2025dcnn}. Collectively, these efforts confirm that EEG carries discriminative information for MCI detection, yet each methodological family carries characteristic limitations that remain unresolved.

Three interrelated limitations persist across these approaches and impede translation into routine clinical practice. First, handcrafted EEG descriptors span heterogeneous physiological domains, including spectral power, entropy measures, Hjorth parameters, sleep spindles, slow oscillations, and demographic covariates, yet conventional classifiers treat this heterogeneous signal as a single concatenated feature vector, discarding the neurophysiologically meaningful domain structure that organizes these features. Second, deep recurrent and convolutional architectures function largely as black-box models, providing negligible transparency into their decision-making processes; in clinical settings where the reasoning behind predictions must be interpretable to establish trust and support informed decision-making, this opacity is a critical barrier. Third, although sleep EEG has been used in prior MCI studies, existing frameworks do not leverage the structured organization of sleep-derived features to explicitly analyze cross-domain interactions, such as the relationship between sleep complexity measures and slow oscillation morphology, that may carry discriminative information beyond individual feature groups.

Concept-based learning offers a principled mechanism for addressing the interpretability and structure limitations by introducing an intermediate representation between raw input features and final predictions~\cite{koh2020concept}. In the EEG-MCI setting, this organization is especially well-motivated because extracted features naturally align with interpretable physiological groups. However, concept-level representations alone do not make cross-concept interactions explicitly analyzable; pairwise relationships among EEG concept groups remain implicit rather than inspectable. Separately, Kolmogorov-Arnold Networks (KANs) have recently emerged as a flexible alternative to conventional multilayer perceptrons by replacing fixed activation functions with learnable univariate functions on network edges, offering nonlinear function approximation potentially well-suited to structured tabular data~\cite{liu2025kan,eslamian2025tabkan}. Yet standard KAN-based tabular classifiers do not encode domain-level physiological concepts or explicitly model higher-order interactions among concept groups. In both the EEG literature and the broader tabular biomedical learning literature, no existing work addresses all three requirements together: preserving physiological concept structure, making cross-concept interactions explicitly analyzable, and performing flexible nonlinear classification in that structured space.

This gap motivates the present work. The field lacks a framework that simultaneously organizes heterogeneous EEG-derived features according to their neurophysiological meaning, expands their pairwise relationships into an inspectable interaction space, and applies nonlinear tabular classification within that structured representation. Addressing this gap requires a design that treats concept structure, interaction modeling, and nonlinear learning as jointly necessary rather than independently sufficient components. This paper proposes Concept-guided Polynomial-transformed Tabular learning using Kolmogorov-Arnold Network (\textit{CPTabKAN}), a framework that fills this gap by learning from physiologically grouped EEG concepts, explicitly expanding their first- and second-order interactions through polynomial feature expansion, and applying a TabKAN classifier in the resulting concept interaction space. The paper then tests whether this joint design improves both prediction and interpretability relative to each component in isolation and relative to a comprehensive set of baselines.

This study pursues three principal objectives: first, to determine whether concept-structured feature organization improves classification performance over flat-feature baselines that operate on unstructured concatenations of the same EEG descriptors; second, to determine whether explicit second-order concept interaction expansion adds measurable value beyond first-order concept representations alone; and third, to assess whether the learned framework surfaces interpretable concept-level and interaction-level importance patterns that are coherent with neurophysiological expectations for MCI. These objectives are evaluated on a real-world sleep EEG cohort under a unified preprocessing and evaluation protocol, with systematic ablation and hyperparameter sensitivity analyses supporting each conclusion.

The contributions of this work are organized along three dimensions. \textbf{Methodologically}, we introduce CPTabKAN, a framework that leverages neurophysiologically organized concept encoders, explicit polynomial concept interaction expansion, and a nonlinear TabKAN classifier for sleep EEG-based MCI detection. To our knowledge, this kind of framework has not been previously explored in EEG-based MCI classification or broader tabular biomedical modeling. \textbf{Empirically}, 
CPTabKAN achieves the best performance among all evaluated traditional, ensemble, and SOTA-inspired deep learning baselines under a unified evaluation protocol. 

\textbf{Interpretively}, we provide concept-level and interaction-level importance analyses that surface which physiological concept groups and their pairwise relationships drive model predictions, offering a transparency mechanism aligned with clinical reasoning about sleep and cognitive health.

\section{Methodology}
\subsection{Study Design Overview}
The study is organized as a subject-level binary classification task. The classification target is cognitive status (MCI versus cognitively normal), derived from MMSE scores recorded at a single clinic visit.

The experimental design comprises four components. First, a comparative evaluation assesses whether CPTabKAN improves classification performance relative to a broad set of classical, ensemble, and deep learning baselines under identical input conditions. Second, a first-order versus second-order comparison isolates the contribution of explicit polynomial concept interactions. Third, a pre-specified ablation study systematically removes individual architectural modules to attribute the observed performance gains to specific components of the framework. Fourth, a concept importance analysis characterizes which concept groups and pairwise interactions the trained model assigns the greatest weight to, providing a structured window into the learned decision logic.

\subsection{Proposed Framework}
\begin{figure}
	\centering
	\includegraphics[width=.95\textwidth]{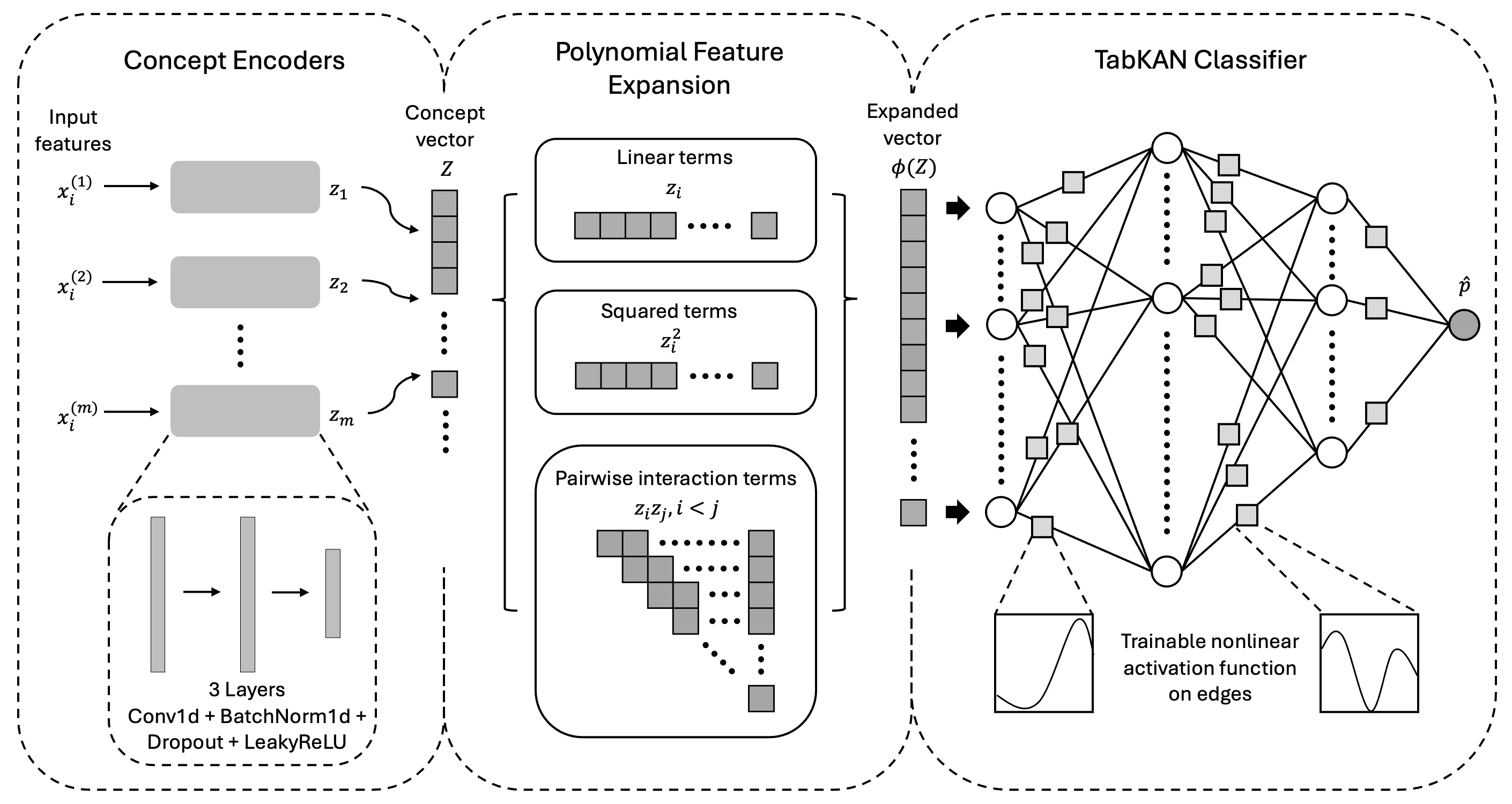}
	\caption{Overview of the CPTabKAN architecture.}
	\label{fig:overview}
\end{figure}

CPTabKAN realizes the subject-level classifier $f_{\theta}: \mathbb{R}^{D}\rightarrow [0,1]$ through three sequentially composed modules, formalized as:
\begin{equation}
\hat{p}_i
=
h_{\theta_h}\!\left(\phi(g_{\theta_g}(x_i))\right).
\label{eq:final}
\end{equation}
where $g_{\theta_g}$ denotes the concept encoders, $\phi(\cdot)$ the polynomial expansion, and $h_{\theta_h}$ the TabKAN classifier. The full model is trained end-to-end using the downstream classification objective. Figure~\ref{fig:overview} illustrates the overall architecture.

\subsubsection{Concept Representation}
The input vector $x_i\in\mathbb{R}^{D}$ is partitioned into $m=10$ domain-informed concept blocks $\{x_i^{(j)}\in\mathbb{R}^{d_j}\}_{j=1}^{10}$, with block dimensions $d_j\in\{3,8,160,160,24,112,640,176,32,64\}$ summing to $D=1,379$. Each block is processed by a dedicated concept encoder $g_j:\mathbb{R}^{d_j}\rightarrow\mathbb{R}$, producing a scalar concept score. The ten scalars are collected into a bottleneck concept vector:
\begin{equation}
z_i = g_{\theta_g}(x_i)
= [g_1(x_i^{(1)}), g_2(x_i^{(2)}), \ldots, g_m(x_i^{(m)})]
\in \mathbb{R}^{m}.
\label{eq:concept_bottleneck}
\end{equation}

Each encoder $g_j$ first projects its input through a linear layer $(d_j\rightarrow64)$ to obtain a 64-dimensional concept embedding $e_j\in\mathbb{R}^{64}$.  The ten embeddings are then concatenated to form a joint representation $h\in\mathbb{R}^{640}$, which is then passed through three sequential grouped $1\times1$ convolutional layers (Conv1d, kernel size 1, groups = 10), reducing the per-group representation from 64 to 64 to 32 dimensions before projecting to a scalar output per concept. LeakyReLU activations and dropout ($p=0.1$) are applied after each hidden layer. The grouped convolution enforces concept-level independence: channels belonging to a given concept group are mixed only within that group.

Compressing heterogeneous, high-dimensional feature blocks into a single scalar per concept is a deliberate design choice to enforce domain-aligned latent structure. Rather than treating all 1,379 features as an undifferentiated input vector, this bottleneck operationalizes the hypothesis that the physiologically meaningful unit of analysis is the concept group, not the individual feature. The encoders are trained exclusively through the downstream task loss without explicit concept supervision; the bottleneck thus imposes domain-informed structure as an inductive bias while allowing the representations to be shaped freely by the classification objective.

\subsubsection{Polynomial Interaction Expansion}
A degree-2 polynomial feature expansion is applied to the concept vector $z_i$, producing all first-order terms, squared terms, and pairwise interaction terms:
\begin{equation}
\phi(z_i) =
\left[
z_i^{(1)}, \ldots, z_i^{(m)},
(z_i^{(1)})^2, \ldots, (z_i^{(m)})^2,
z_i^{(1)}z_i^{(2)}, \ldots, z_i^{(m-1)}z_i^{(m)}
\right].
\end{equation}
For $m=10$, this yields $\binom{m+2}{2}-1 = 66$ dimensions (bias term excluded).

The polynomial expansion makes cross-concept effects explicit and directly testable. Rather than allowing a downstream network to discover interactions implicitly through depth, this module encodes pairwise concept relationships as observable input features, enabling both the classifier and the importance analysis to operate on named, interpretable terms. The degree is fixed at 2 to balance representational expressiveness with interpretability; higher-order expansions would increase dimensionality combinatorially and obscure concept-level attribution.

\subsubsection{Nonlinear Classification}
The expanded representation $\phi(z_i)\in\mathbb{R}^{66}$ is passed to a TabKAN classifier $h_{\theta_h}$ \cite{eslamian2025tabkan}, grounded in Kolmogorov-Arnold Networks (KANs) \cite{liu2025kan}. KANs replace fixed nonlinear activations on neurons with learnable univariate functions on network edges, enabling flexible function approximation with improved interpretability relative to standard multilayer perceptrons. In the present implementation, edge functions are parameterized as truncated Fourier series using FourierKAN basis functions \cite{xu2025enhancing}, the native parameterization of this architecture, rather than the B-splines used in the original KAN formulation. FourierKAN was selected based on superior empirical performance on the present dataset relative to B-spline alternatives. The TabKAN classifier uses two layers: the first layer has 64 neurons with grid size 6, and the second has 512 neurons with grid size 7; these hyperparameters were selected via grid search (Appendix~\ref{appendix:hyperparameter-selection}).  Furthermore, sensitivity analyses with respect to key CPTabKAN hyperparameters are reported in Appendix~\ref{appendix:hyperparameter-sensitivity}.

\subsubsection{Training Objective}
The full model is trained end-to-end by minimizing:
\begin{equation}
\mathcal{L}=\mathcal{L}_{CE}(\hat{y},y)+\lambda\cdot mean(z^2)
\label{eq:loss}
\end{equation}
where $\mathcal{L}_{CE}$ is the cross-entropy loss, $z\in\mathbb{R}^{10}$ is the concept bottleneck vector from Eq. (\ref{eq:concept_bottleneck}), and $\lambda=0.05$ is a fixed regularization coefficient.  The second term applies an L2-style shrinkage penalty on the concept representations to prevent magnitude inflation and improve generalization.  Optimization details are provided in Appendix~\ref{appendix:optimization_details}.

\subsection{Data Source and Cohort}
The Study of Osteoporotic Fractures (SOF) is a longitudinal cohort study that initially enrolled 9,704 Caucasian women aged 65 years and older from four U.S. metropolitan centers between 1986 and 1988, with an additional 662 African-American women recruited between 1997 and 1998 \cite{spira2008sleep}. The present analysis draws on data from the eighth clinic visit (2002–2004), during which 461 women participated in overnight in-home polysomnography (PSG). Of these, 438 also completed standardized cognitive assessments; the remaining 23 were excluded due to missing MMSE scores. After further exclusion of subjects with incomplete data across all extracted features, 372 subjects were retained as the analytic sample. PSG recordings were acquired using the Compumedics Profusion system and captured EEG, electrooculographic (EOG), and electromyographic (EMG) signals in 30-second epochs stored in European Data Format (EDF). Recording files and cognitive assessment scores were accessed through the National Sleep Research Resource (NSRR) \cite{zhang2018national, nsrr} and the SOF Online repository \cite{sof}.

The analytic cohort had a mean age of 82.7 years, $93.6\%$ were Caucasian, and mean years of education was 12.5. Cognitive status was classified using MMSE scores recorded at the eighth clinic visit only: subjects with MMSE below 27 were classified as MCI-positive ($y_i=1$); all others were classified as cognitively normal ($y_i=0$) following established thresholds in the EEG-based MCI literature \cite{geng2022sleep, Aljalal2024, diagnostics14151619}. This yielded 42 MCI and 330 cognitively normal subjects, corresponding to a class imbalance ratio of approximately 7.9:1. The SOF dataset represents one of the largest publicly available sleep EEG repositories with accompanying cognitive assessments, providing substantially greater statistical power than prior EEG-based MCI studies, which have typically relied on cohorts of 23–61 subjects.

\subsection{Feature Construction and Concept Grouping}
A multi-domain feature extraction pipeline was applied to eight EEG channels: two central electrodes (C3, C4), two mastoid electrodes (M1, M2), and four derived channels (C3-M2, C4-M1, C3-LM, C4-LM) obtained via contralateral and linked mastoid referencing schemes \cite{Purcell2017, kozhemiako2022sources}. The ten concept groups and their physiological rationale are summarized in Table~\ref{tbl:dataset}, with extractor-level implementation details provided in Appendix~\ref{appendix:feature_extraction_detail}.  EEG raw data preprocessing followed a standardized pipeline, with comprehensive details reported in Appendix~\ref{appendix:preprocessing}.

\begin{table}[width=1\linewidth,cols=4,pos=h]
\caption{High-level concept groups and corresponding feature categories.  All per-channel features are computed across two central electrodes, two mastoid electrodes, and four derived EEG channels; demographic and SVD features are subject-level or cross-channel.}\label{tbl:dataset}
\begin{tabular*}{\tblwidth}{@{} CLCL@{} }
\toprule
\begin{tabular}{@{}c@{}} Concept \\ Index \end{tabular} & High-Level Concept & \begin{tabular}{@{}c@{}} No. of \\ Features \end{tabular} & Low-Level Feature Description \\
\midrule
$z_1$ & Demographics & 3 & Age, years of education, ethnicity \\
$z_2$ & Lempel-Ziv-Welch (LZW) & 8 & LZW compression index (per channel) \\
$z_3$ & Multi-Scale Entropy (MSE) & 160 & MSE across 20 temporal scales (per channel) \\
$z_4$ & Power Spectral Density (PSD) & 160 & Absolute and relative spectral band powers (per channel) \\
$z_5$ & Hjorth Parameters & 24 & Activity, mobility, complexity (per channel) \\
$z_6$ & Slow Oscillations (SO) & 112 & Rate, amplitude, duration, slope morphology (per channel) \\
$z_7$ & Spindles & 640 & Density, amplitude, duration, chirp, symmetry (per channel) \\
$z_8$ & Statistics & 176 & Temporal signal statistics (per channel) \\
$z_9$ & Singular Value Decomposition (SVD) & 32 & Singular values, explained variance (cross-channel) \\
$z_{10}$ & SVD Component Weights & 64 & Channel-level component contributions (cross-channel) \\
\bottomrule
\end{tabular*}
\end{table}

\subsection{Baseline Comparison Strategy}
To contextualize CPTabKAN's performance, we implemented a comprehensive suite of baseline classifiers representing distinct machine learning paradigms.  Direct reproduction of published state-of-the-art (SOTA) methods was not feasible due to unavailable source code, inaccessible datasets, insufficient implementation details, and differing experimental protocols.  Instead, we reimplemented the core architectural principles of each SOTA method under a unified preprocessing, feature representation, and evaluation framework, enabling fair relative comparison.

The implemented baselines include classifier-level or architecture-inspired reimplementations corresponding to methods used in prior EEG-based MCI studies, including SVM-based pipelines~\cite{geng2022sleep, Aljalal2024}, a KNN-based pipeline~\cite{diagnostics14151619}, Random Forest-based pipelines~\cite{rutkowski2024mild}, Bi-LSTM and Bi-GRU architectures~\cite{geng2022sleep,Said2024}, and the DCNN-SBiL architecture~\cite{devi2025dcnn}. We also implemented additional commonly used tabular-learning baselines, including ExtraTrees, XGBoost, LightGBM, GradientBoosting, and Logistic Regression.

All recurrent baselines (Bi-GRU, Bi-LSTM) and DCNN-SBiL were applied to the aggregated subject-level tabular feature vector rather than raw EEG epoch sequences.  Following feature extraction and epoch aggregation, each subject was represented by a single fixed-length feature vector of dimensionality equal to the total number of extracted features.  This representation was used consistently across all models, both baselines and the proposed CPTabKAN, to ensure a fair comparison under identical input conditions.  Consequently, recurrent architectures were not used to model temporal dependencies across raw epochs; rather, they served as nonlinear classifiers on the same tabular representation, enabling direct evaluation of differences in model learning capacity while controlling for input representation.

Computational environment details, including hardware, software versions, and library dependencies, and more baseline model implementation details are provided in Appendix~\ref{appendix:computational_environment}.

\subsection{Evaluation Protocol}
\textbf{Primary and secondary settings.} The no-SMOTE condition, in which all models are trained on the original imbalanced data, is the primary evaluation setting. The SMOTE condition, in which synthetic minority-class samples are generated to achieve full 1:1 class balancing, is included as a secondary robustness analysis. This ordering reflects the view that the imbalanced setting is more representative of real-world deployment conditions, and that SMOTE results provide supporting, rather than co-primary, evidence. SMOTE robustness results are reported in Appendix~\ref{appendix:smote}.

\textbf{Cross-validation.} Model performance was evaluated using 10-fold cross-validation with shuffling. No held-out test set was used; all performance estimates are derived from cross-validation. For the SMOTE condition, synthetic samples were generated exclusively within each training partition after splitting, preventing synthetic data from appearing in any evaluation fold and thus avoiding information leakage. Full cross-validation and class-balancing implementation details are provided in Appendix~\ref{appendix:CV-ClassBalancing}.

\textbf{Primary metric.} The primary metric is the mean weighted $F1$-score across folds (mean $\pm$ std), selected for its robustness to the 7.9:1 class imbalance \cite{Salmi2024, app14219863, Hancock2023}. Moreover, weighted precision and recall are reported as secondary descriptive metrics.

\textbf{Statistical comparison.} To assess whether the performance advantage of CPTabKAN-Second Order over the strongest baseline was consistent across folds, a one-sided paired $t$-test was applied to the 10 per-fold weighted $F1$-scores under the directional hypothesis $H_1: \mu_{diff} > 0$. The comparison target (GradientBoosting, as the strongest classical baseline) and the significance threshold ($\alpha=0.05$) were pre-specified before results were examined. No correction for multiple comparisons was applied, as the statistical inference was restricted to this single pre-specified primary comparison.

\textbf{Ablation logic.} The ablation study is pre-specified as part of the study design. It evaluates six configurations obtained by systematically removing one or more of the three modules, Concept Encoders (CE), Polynomial Feature Expansion (PFE), and TabKAN classifier, to isolate the independent contribution of each component. The ablation is conducted under the primary (no-SMOTE) condition only.

\subsection{Concept Importance Analysis}
Concept importance scores are estimated from the trained CPTabKAN-Second Order model by aggregating the absolute magnitudes of the FourierKAN edge coefficients associated with each first-order concept term and each second-order pairwise interaction term. This operationalization is the architecture's native parameterization: because FourierKAN represents each edge transformation as a truncated Fourier series, the coefficient magnitudes directly reflect the learned strength of each edge's functional contribution, and gradient-based alternatives were not pursued. The resulting scores characterize which concept groups and concept interactions the trained model assigns the greatest weight to.

\section{Results}
\subsection{Performance Relative to Flat-Feature Baselines}
\textit{Does the proposed framework improve MCI classification over established tabular and deep baselines under the primary evaluation setting?}

Table~\ref{tbl:result} presents the mean ± standard deviation of weighted $F1$-score, precision, and recall for all evaluated models under both experimental conditions. Under the primary (no-SMOTE) condition, CPTabKAN-Second Order achieved a mean weighted $F1$-score of $0.9038 \pm 0.034$, representing an absolute improvement of 5.65 percentage points over GradientBoosting ($0.8473 \pm 0.080$), the strongest baseline. CPTabKAN-First Order achieved $0.9007 \pm 0.053$, an improvement of 5.34 percentage points over the same comparator. Among the SOTA-inspired deep reimplementations, Bi-GRU achieved the highest $F1$-score ($0.8452 \pm 0.048$), followed by DCNN-SBiL ($0.8432 \pm 0.057$) and Bi-LSTM ($0.8420 \pm 0.050$); classical ensemble methods clustered in the range $0.8337-0.8473$. These results indicate that strong classical and ensemble tabular learners do not already solve the task at the level achieved by the proposed framework, and that deep nonlinear capacity alone, as represented by the reimplemented recurrent and convolutional baselines under the same tabular input, does not account for the observed gains. A one-sided paired $t$-test on the 10 per-fold $F1$-scores of CPTabKAN-Second Order versus GradientBoosting yielded $t(9)=1.934, p=0.043$ under the pre-specified directional hypothesis, providing fold-level statistical support for the performance advantage. Given the limited number of folds and the absence of an external test cohort, this result should be interpreted as supportive rather than definitive evidence.

CPTabKAN-Second Order exhibited the lowest standard deviation among all deep learning methods ($\sigma = 0.034$ without SMOTE), indicating stable generalization across folds. This pattern of both higher mean performance and lower fold-to-fold variance supports the proposed design hypothesis that concept-structured representation and interaction-aware modeling produce a more consistent decision boundary than unstructured alternatives. Precision and recall results are consistent with the $F1$ pattern and are provided in Table~\ref{tbl:result} for completeness. Performance advantages persisted under SMOTE-based class balancing (Appendix~\ref{appendix:smote}).

\begin{table}[width=.95\linewidth,cols=7,pos=h]
\caption{Comparative classification performance of all evaluated models on the SOF dataset under 10-fold cross-validation.  Results are reported as mean $\pm$ standard deviation of the weighted $F1$-score (F1), weighted precision (P), and weighted recall (R) across 10 folds.  Superscript $\dagger$ denotes SOTA-inspired reimplementations; superscript $\ddagger$ denotes additional baseline algorithms. Bold values indicate the best performance within each condition, and underlined values indicate the second-best performance.
}\label{tbl:result}
\begin{tabular*}{\tblwidth}{@{\extracolsep\fill}lcccccc}
\toprule%
& \multicolumn{3}{@{}c@{}}{Without Class Balancing} & \multicolumn{3}{@{}c@{}}{With Class Balancing} \\\cmidrule{2-4}\cmidrule{5-7}%
Algorithm & F1 (mean $\pm$ std) $\uparrow$ & P $\uparrow$ & R $\uparrow$ & F1 (mean $\pm$ std) $\uparrow$ & P $\uparrow$ & R $\uparrow$ \\
\midrule
Bi-GRU \cite{geng2022sleep}$^\dagger$ & 0.8452 $\pm$ 0.048 & 0.8188 & 0.8923 & 0.6984 $\pm$ 0.210 & 0.7898 & 0.6891\\
SVM \cite{geng2022sleep, Aljalal2024}$^\dagger$ & 0.8337 $\pm$ 0.088 & 0.7899 & 0.8844 & 0.7942 $\pm$ 0.076 & 0.8420 & 0.7664\\
Bi-LSTM \cite{Said2024}$^\dagger$ & 0.8420 $\pm$ 0.050 & 0.8088 & 0.8869 & 0.7741 $\pm$ 0.075 & 0.8014 & 0.7610\\
KNN \cite{diagnostics14151619}$^\dagger$ & 0.8351 $\pm$ 0.087 & 0.7903 & 0.8871 & 0.6747 $\pm$ 0.075 & 0.8550 & 0.6000\\
RandomForest \cite{rutkowski2024mild}$^\dagger$ & 0.8337 $\pm$ 0.087 & 0.7900 & 0.8844 & 0.8317 $\pm$ 0.102 & 0.8432 & 0.8445\\
ExtremeRandomTrees$^\ddagger$ & 0.8337 $\pm$ 0.088 & 0.7899 & 0.8844 & 0.8401 $\pm$ 0.098 & 0.8023 & 0.8845\\
LogisticRegression$^\ddagger$ & 0.8464 $\pm$ 0.099 & 0.8158 & 0.8845 & 0.8074 $\pm$ 0.070 & 0.8494 & 0.7824\\
GradientBoosting$^\ddagger$ & 0.8473 $\pm$ 0.080 & 0.8309 & 0.8870 & 0.8464 $\pm$ 0.092 & 0.8318 & 0.8710\\
LightGBM$^\ddagger$ & 0.8444 $\pm$ 0.098 & 0.8061 & 0.8925 & 0.8424 $\pm$ 0.107 & 0.8249 & 0.8683\\
XGBoostClassifier$^\ddagger$ & 0.8351 $\pm$ 0.087 & 0.7903 & 0.8871 & 0.8437 $\pm$ 0.093 & 0.8290 & 0.8737\\
VotingEnsemble $^\ddagger$ & 0.8351 $\pm$ 0.087 & 0.7903 & 0.8871 & 0.8466 $\pm$ 0.104 & 0.8174 & 0.8818\\
DCNN-SBiL \cite{devi2025dcnn}$^\dagger$ & 0.8432 $\pm$ 0.057 & 0.8069 & 0.8897 & 0.7661 $\pm$ 0.098 & 0.8147 & 0.7507\\
\midrule
CPTabKAN-First Order & \underline{{0.9007 $\pm$ 0.053}} & \textbf{0.9245} & \textbf{0.9166} & \underline{{0.8735 $\pm$ 0.024}} & \underline{{0.8849}} & \underline{{0.8869}}\\
CPTabKAN-Second Order & \textbf{0.9038 $\pm$ 0.034} & \underline{{0.9177}} & \underline{{0.9139}} & \textbf{0.8902 $\pm$ 0.046} & \textbf{0.8869} & \textbf{0.9030}\\
\bottomrule
\end{tabular*}
\end{table}

\subsection{Contribution of Explicit Concept Interactions}
\label{sec:interactions}
\textit{Does making cross-concept interactions explicit through polynomial expansion add discriminative value beyond first-order concept scores?}

CPTabKAN-Second Order ($F1=0.9038$) outperformed CPTabKAN-First Order ($F1=0.9007$) by $\Delta F1=0.0031$ under the primary condition. This incremental but consistent gain, observed despite the modest magnitude, is consistent with the hypothesis that second-order concept interactions encode subtle joint dependencies not captured by individual concept scores. The relative importance of specific interactions is analyzed in Section~\ref{sec:importance}.

\subsection{Ablation Evidence for the Source of Performance Gains}
\label{sec:ablation}
\textit{Can the performance improvement be attributed to a single generic capacity effect, or do all three architectural modules contribute independently?}

To isolate the contribution of each module, six configurations were evaluated by systematically removing CE, PFE, and the TabKAN classifier in combination. Removing CE while retaining the remaining modules yielded $F1 = 0.8496$, a degradation of 5.42 percentage points from the full model; removing both CE and PFE left TabKAN on unstructured features at $F1 = 0.8489$, confirming that polynomial expansion over raw inputs provides no benefit without prior concept-level compression. Replacing TabKAN with the best alternative classifier while retaining CE and PFE yielded $F1 = 0.8748$, a 2.90-point reduction, and removing PFE alone (CPTabKAN-First Order) produced an incremental decrease of 0.0031. These results confirm that all three modules contribute independently; no single component accounts for the full performance gain, and the concept encoders provide the largest individual contribution. Ablation study details are provided in Appendix~\ref{appendix:ablation}.

\subsection{Concept-Level Interpretability Findings}
\label{sec:importance}
\textit{Which concept groups and concept interactions does the trained model assign the greatest weight to, and what does this reveal about the learned decision structure?}

\begin{figure}
	\centering
	\includegraphics[width=.9\textwidth]{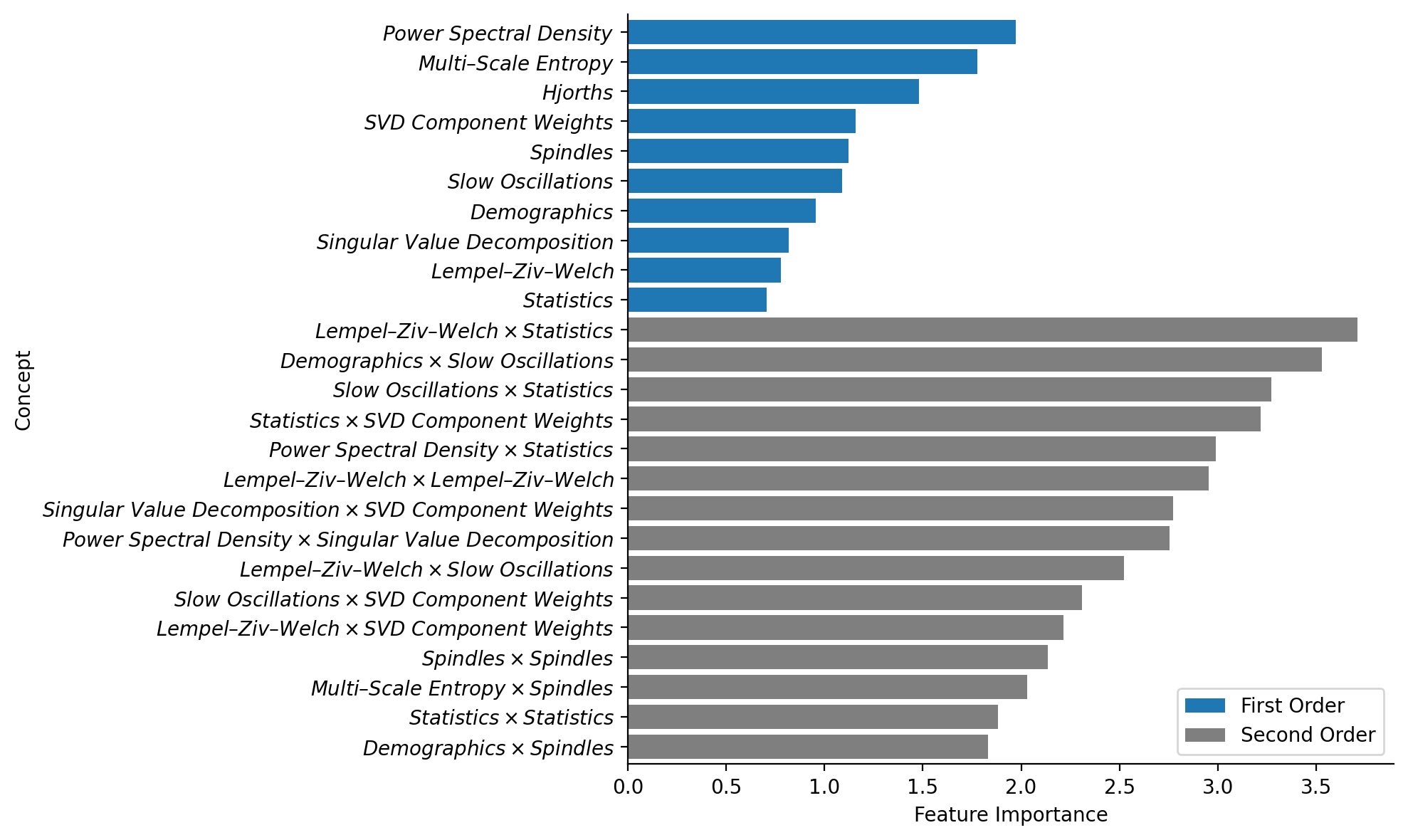}
	\caption{Concept importance scores for CPTabKAN-Second Order, derived from absolute Fourier-series KAN coefficient magnitudes. Upper panel: all 10 first-order concept importances. Lower panel: top 15 second-order pairwise interaction importances. The larger magnitudes of several second-order terms suggest that CPTabKAN assigns substantial model capacity to cross-concept interactions, although these scores should be interpreted as model-derived importance proxies rather than causal effects.}
	\label{fig:featureimportance}
\end{figure}

Figure~\ref{fig:featureimportance} presents the concept importance scores derived from CPTabKAN-Second Order, computed as the absolute magnitudes of the FourierKAN edge coefficients, the native parameterization of this architecture, associated with each first-order and second-order term. These scores reflect the learned importance structure of the model and should be interpreted as model-derived association measures rather than causal biomarkers; they indicate which concept groups and concept interactions the trained model relies on for prediction, but they do not establish mechanistic or clinical causality.

\textbf{First-order importance pattern.} Among the ten first-order concepts, Power Spectral Density (PSD; importance = 1.974) and Multi-Scale Entropy (MSE; importance = 1.779) exhibited the highest individual contributions, followed by Hjorth parameters (1.481), SVD Component Weights (1.160), and Spindles (1.122). Slow Oscillations (1.089), Demographics (0.955), SVD (0.816), LZW (0.779), and Statistics (0.705) contributed in the lower tier, with all concepts registering non-negligible scores. This pattern is consistent with the intuition that MCI-related changes in sleep EEG may affect oscillatory power distributions, signal complexity, and time-domain dynamics in a complementary, multi-domain fashion; CPTabKAN appears to leverage information across all ten concept groups rather than concentrating on a single descriptor family.

\textbf{Second-order importance pattern.} Among the top-15 pairwise interaction terms, LZW $\times$ Statistics achieved the highest importance score (3.709), followed by Demographics $\times$ Slow Oscillations (3.528), Slow Oscillations $\times$ Statistics (3.272), Statistics $\times$ SVD Component Weights (3.217), and PSD $\times$ Statistics (2.989). All top-15 pairs yielded scores exceeding 1.83, and the top-5 pairs each exceeded 2.99, substantially above the highest first-order score of 1.974. 
The larger magnitudes of several second-order terms suggest that the trained model assigns substantial functional weight to cross-concept relationships, although these scores should be interpreted as model-derived importance proxies rather than directly comparable causal effects. The prominence of interactions involving LZW, Statistics, Demographics, and Slow Oscillations suggests that algorithmic complexity measures may be most informative when interpreted jointly with basic waveform distribution characteristics, and that age- and education-related factors may modulate the relationship between sleep oscillatory dynamics and cognitive status in ways that neither concept captures independently. Combined with the $\Delta F1=0.0031$ improvement from adding the polynomial expansion, these findings support the conclusion that second-order concept interactions encode discriminative information beyond first-order concept representations.

\section{Discussion}

The central finding of this study is that a framework accounting for neurophysiological concept structure, making cross-concept interactions explicitly analyzable, and applying flexible nonlinear tabular classification in the resulting space yields a measurable and stable improvement in sleep EEG-based MCI classification. CPTabKAN-Second Order achieved a mean weighted $F1$-score of $0.9038 \pm 0.034$ under the primary no-SMOTE condition, outperforming the strongest baseline, GradientBoosting, by 5.65 percentage points ($t(9)=1.934$, $p=0.043$, one-sided paired $t$-test), and exhibited the lowest fold-to-fold variance among all evaluated deep learning configurations. Critically, this result was obtained without any single-module explanation: the ablation evidence confirms that neither concept structuring, interaction modeling, nor nonlinear classification alone accounts for the performance, and must instead be understood as the joint product of all three. In doing so, CPTabKAN simultaneously addresses the three interrelated limitations identified in the Introduction, the flat-vector treatment of heterogeneous feature families, the opacity of existing deep architectures, and the absence of an inspectable cross-domain interaction mechanism, none of which were resolved by any prior EEG-MCI framework individually.

Prior EEG-based MCI studies have established predictive feasibility across SVM, KNN, random forest, recurrent, and hybrid architectures \cite{geng2022sleep,Aljalal2024,diagnostics14151619,rutkowski2024mild,Said2024,devi2025dcnn}, yet each framework treats the extracted EEG descriptor set as a single concatenated vector, leaving physiological feature-family structure implicit. The present results suggest that this design choice is not merely an aesthetic limitation: under identical input conditions, models that do not preserve concept organization, including strong nonlinear deep baselines such as Bi-GRU ($F1=0.8452$), Bi-LSTM ($F1=0.8420$), and DCNN-SBiL ($F1=0.8432$), failed to reach the performance level of CPTabKAN. The gap cannot be attributed to representational capacity differences between deep and shallow models per se, because these recurrent and convolutional baselines already employ substantial nonlinear capacity on the same tabular input. Instead, the results support the interpretation that preserving the physiologically meaningful grouping of spectral, entropy, thalamocortical, and demographic features as distinct inductive units provides a more informative inductive bias than treating the same features as an undifferentiated 1,379-dimensional vector.

The concept importance findings provide a structured window into the model's decision logic, while also clarifying what such findings can and cannot claim. At the first-order level, the dominance of PSD (importance $= 1.974$), MSE ($1.779$), and Hjorth parameters ($1.481$) is broadly consistent with established neurophysiological evidence that MCI-related changes in sleep EEG affect oscillatory power distributions and signal complexity \cite{geng2022sleep,Aljalal2024}: PSD captures frequency-band power shifts sensitive to neurodegenerative changes \cite{li2024data}, while MSE characterizes temporal irregularity across multiple scales in a manner that is complementary to spectral descriptors. The non-negligible contributions from sleep spindle and slow oscillation concepts are coherent with evidence that thalamocortical and corticohippocampal slow-wave events are implicated in memory consolidation and are altered in early cognitive decline \cite{wei2018differential,ng2025bayesian,Ladenbauer2017zu}. At the second-order level, the largest interaction weights, LZW $\times$ Statistics ($3.709$), Demographics $\times$ Slow Oscillations ($3.528$), Slow Oscillations $\times$ Statistics ($3.272$), and PSD $\times$ Statistics ($2.989$), exceed the highest first-order score, suggesting that the trained model assigns substantial functional weight to cross-concept relationships. The prominence of demographics interacting with slow oscillation structure is plausibly consistent with evidence that age and education modulate sleep oscillatory dynamics in the aging brain \cite{campos2026differences,niethard2023aging,zust2023hierarchy}, and that these demographic factors may therefore condition the relationship between slow-wave activity and cognitive status in ways that neither concept captures independently. However, these importance scores are coefficient-based, model-derived association measures from a single cross-sectional cohort; they indicate which concept groups and interactions the trained model relies on for prediction but do not establish mechanistic, causal, diagnostic, or biomarker-level relationships between EEG feature families and MCI pathophysiology.

Several limitations constrain the conclusions that can be drawn from this study. First, the analysis is based on a single cohort from the Study of Osteoporotic Fractures (SOF). Although SOF provides a relatively large sleep EEG dataset compared with prior EEG-based MCI studies, the cohort primarily consists of elderly Caucasian women and is not demographically representative of the broader aging population. The generalizability of CPTabKAN to men, younger elderly adults, racially and ethnically diverse cohorts, and clinical populations with different comorbidity profiles cannot be established from these results. Second, MCI labels were derived using an MMSE threshold rather than comprehensive clinical diagnosis; MMSE-based labeling provides a practical and reproducible criterion but may introduce label noise, particularly for borderline cases or individuals whose cognitive impairment is not fully captured by global cognitive screening. Third, the absence of an independent external validation cohort means that all performance estimates are internal cross-validation results, which assess consistency of the framework within SOF but do not establish transportability across cohorts, acquisition systems, scoring protocols, or clinical settings. Fourth, although CPTabKAN improves interpretability through concept-level organization and coefficient-based importance analysis, the concept bottleneck representations are not explicitly supervised against clinical concepts; they should therefore be understood as domain-structured latent encodings shaped by the classification objective, not as validated clinical biomarkers. 

Future work should pursue external validation on independent sleep EEG cohorts to assess robustness across populations and recording systems. Incorporating epoch-level temporal representations alongside aggregated subject-level features may recover temporal dynamics that are lost during the current feature aggregation pipeline. Longitudinal analyses, particularly prediction of conversion from cognitively normal status to MCI or from MCI to dementia, would leverage the sequential structure of cohorts such as SOF and provide more clinically specific modeling targets than cross-sectional cognitive status classification. Richer clinical labels based on comprehensive neuropsychological assessment, consensus diagnosis, or biomarker-confirmed subtype would also strengthen the clinical validity of future modeling efforts beyond what MMSE-threshold classification permits.

\section{Conclusion}
This study demonstrates that our framework for preserving neurophysiological concept structure, making cross-concept interactions explicitly analyzable through polynomial expansion, and applying flexible nonlinear tabular classification in the resulting structured space yields measurable and stable improvements in sleep EEG-based MCI classification. Ablation confirms that all three modules contribute independently, and their combination outperformed baselines consistently across folds and under class-balancing interventions. The concept-level and interaction-level importance analyses further indicate that the framework surfaces decision structure coherent with neurophysiological expectations for MCI-related sleep alterations, providing a transparency mechanism that neither flat-feature classifiers nor unstructured deep models can natively offer.

These findings support concept-structured, interaction-aware tabular learning as a promising paradigm for interpretable biomedical classification from physiologically organized feature spaces. The conclusions are, however, bounded to the Study of Osteoporotic Fractures cohort, and external validation across demographically diverse populations, independent recording systems, and richer clinical labels remains an essential step before any generalization claim can be made. More broadly, CPTabKAN encodes a design principle, preserving domain structure as an inductive bias, that warrants evaluation in other heterogeneous biomedical tabular settings.

\section*{Declaration of competing interest}
The authors declare that they have no known competing financial interests or personal relationships that could have appeared to influence the work reported in this paper.

\section*{Declaration of generative AI and AI-assisted technologies in the manuscript preparation process}
During the preparation of this work the authors used ChatGPT from OpenAI in order to check the grammar and improve the clarity and readability of the paper.  After using this tool/service, the authors reviewed and edited the content as needed and take full responsibility for the content of the published article.

\section*{Data availability}
All data that support the findings of this study are referenced in this paper.

\section*{Acknowledgments}
This work was supported in part by the NSF under Grants IIS 2327113 and ITE 2433190; and the NIH under Grants P30AG072946. We also acknowledge NSF-supported AI research resources through NAIRR240219, including Jetstream2 and PSC, as well as the University of Kentucky Center for Computational Sciences and Information Technology Services Research Computing, for providing computational support and access to the LCC. The Study of Osteoporotic Fractures (SOF) was supported by National Institutes of Health grants (AG021918, AG026720, AG05394, AG05407, AG08415, AR35582, AR35583, AR35584, RO1 AG005407, R01 AG027576-22, 2 R01 AG005394-22A1, 2 RO1 AG027574-22A1, HL40489, T32 AG000212-14).  The National Sleep Research Resource was supported by the National Heart, Lung, and Blood Institute (R24 HL114473, 75N92019R002).  We gratefully acknowledge Nataliia Kozhemiako of Brigham and Women's Hospital, Harvard Medical School, for generously sharing her EEG preprocessing script.  Her contribution substantially enhanced the efficiency of our research and facilitated timely completion of the study.

\printcredits

\appendix 
\setcounter{table}{0}
\setcounter{figure}{0}
\setcounter{equation}{0}
\renewcommand{\thetable}{\thesection.\arabic{table}}
\renewcommand{\thefigure}{\thesection.\arabic{figure}}
\renewcommand{\theequation}{\thesection.\arabic{equation}}
\counterwithin{figure}{section}
\section*{Appendix}

\section{Hyperparameter Selection}
\label{appendix:hyperparameter-selection}
For CPTabKAN, hyperparameter selection was performed via grid search over TabKAN-specific parameters, including the number of layers, neurons per layer, and FourierKAN grid size.  All 10-fold configurations were evaluated for each hyperparameter setting, and the configuration yielding the highest mean weighted $F1$-score across folds was selected.  Baseline model hyperparameters were selected analogously via Bayesian search, with the best configuration determined by the highest mean weighted $F1$-score across the same 10 folds.

\section{Hyperparameter Sensitivity}
\label{appendix:hyperparameter-sensitivity}
\begin{figure}
	\centering
	\includegraphics[width=.9\textwidth]{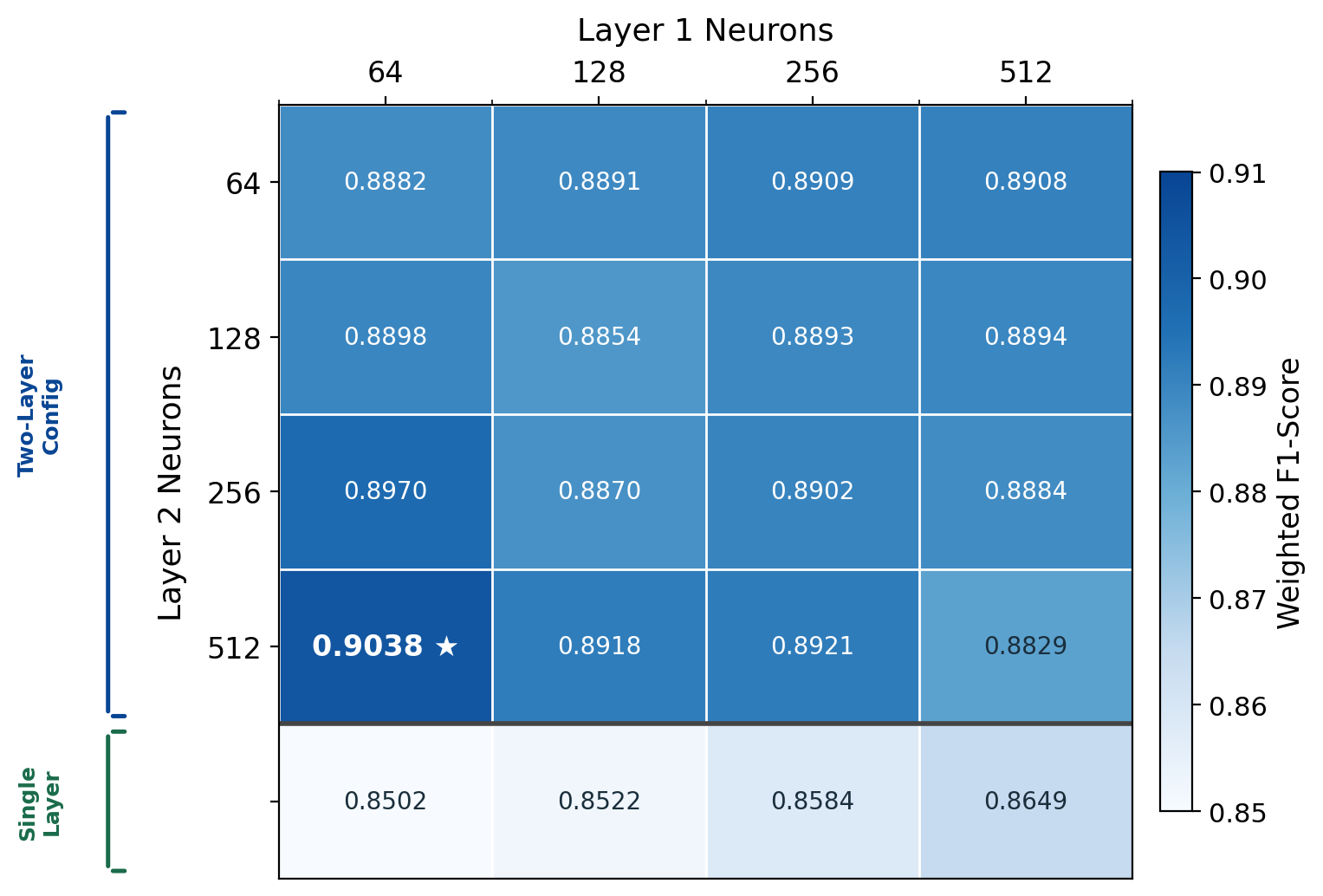}
	\caption{Heatmap of maximum weighted $F1$-score for each two-layer CPTabKAN neuron configuration, maximized over all grid size combinations.  Columns represent layer 1 neuron count; rows represent layer 2 neuron count.  The bottom row (Single-Layer) shows the single-layer baseline for reference; the horizontal line separates single-layer from two-layer results. The bold cell ($\bigstar$) indicates the best-performing configuration within the explored search space (Layer 1 Neurons = 64, Layer 2 Neurons = 512; $F1 = 0.9038$).}
	\label{fig:heatmapneurons}
\end{figure}

\begin{figure}
	\centering
	\includegraphics[width=.95\textwidth]{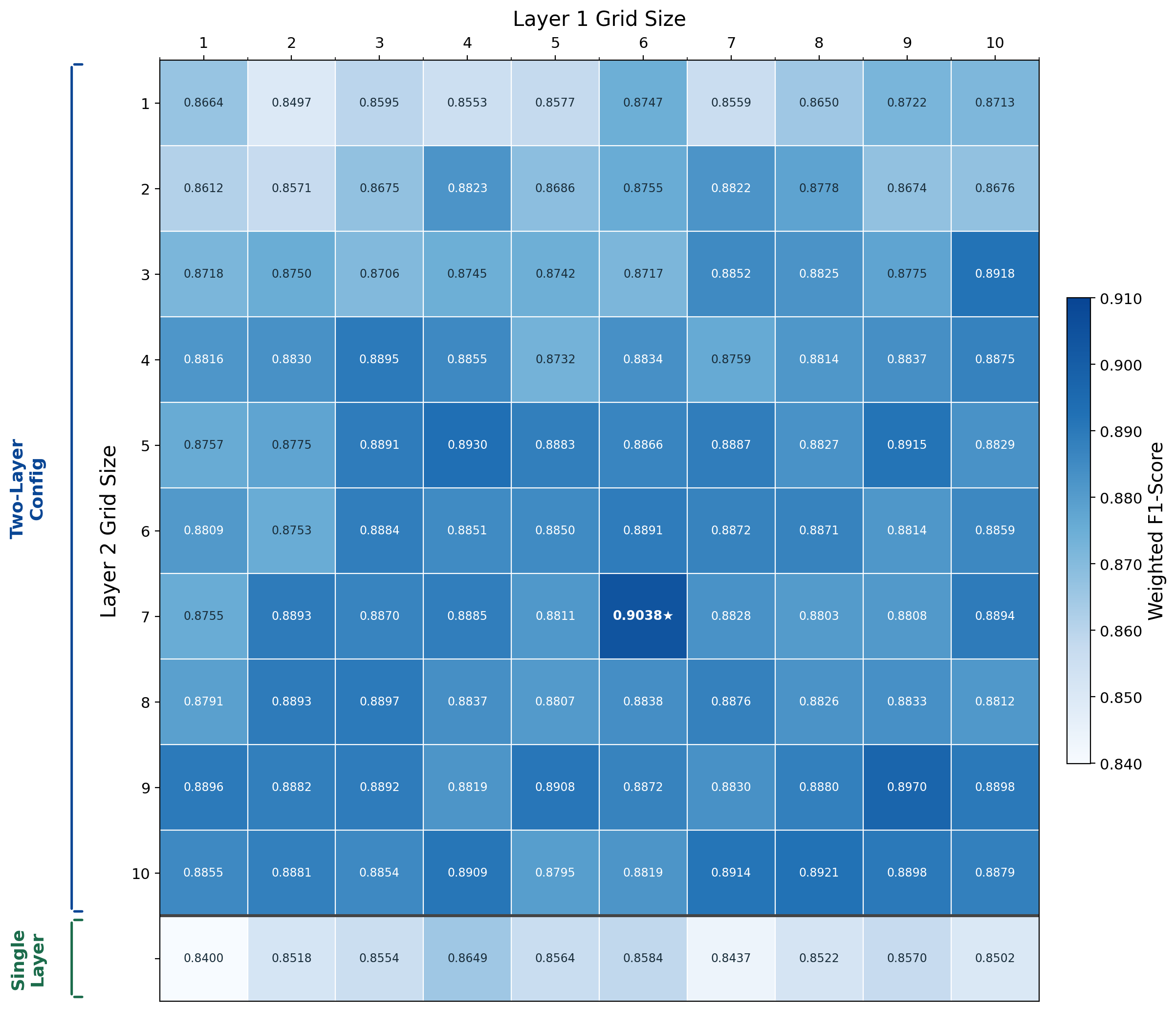}
	\caption{Heatmap of maximum weighted $F1$-score for each two-layer CPTabKAN grid size configuration, maximized over all neuron count combinations.  Columns represent layer 1 grid size; rows represent layer 2 grid size.  The bottom row (Single-Layer) shows the single-layer baseline across grid sizes for reference.  
    The bold cell ($\bigstar$) indicates the best-performing configuration within the explored search space (Layer 1 grid size = 6, Layer 2 grid size = 7; $F1 = 0.9038$).}
	\label{fig:heatmapgrids}
\end{figure}

The sensitivity of CPTabKAN to architectural hyperparameters was evaluated across two configurations, a single-layer and a two-layer TabKAN stack, by exhaustively varying the number of neurons per layer $\in \{64, 128, 256, 512\}$ and the grid size per layer $\in \{1, 2, \ldots, 10\}$.  For each unique combination of neuron counts (\autoref{fig:heatmapneurons}) and grid sizes (\autoref{fig:heatmapgrids}), the maximum weighted $F1$-score across all other hyperparameter combinations is reported, enabling a focused comparison of the contribution of each dimension independently.

\textbf{Effect of Neuron Configuration.}  \autoref{fig:heatmapneurons} presents the maximum weighted $F1$-score for each combination of layer 1 and layer 2 neuron counts, with the single-layer baseline displayed in the bottom row for direct comparison.  Single-layer configurations achieved $F1$-scores ranging from $0.8502$ (64 neurons) to $0.8649$ (512 neurons), consistently below all two-layer configurations irrespective of neuron count.  Among two-layer configurations, performance was generally robust across layer 1 neuron choices when layer 2 neurons were set to 512, with layer 1 neurons = 64, layer 2 neurons = 512 achieving the best performance within the explored search space, with $F1 = 0.9038$, an improvement of $3.89$ percentage points over the best single-layer result.  The second-best configuration (layer 1 neurons = 64, layer 2 neurons = 256; $F1 = 0.8970$) confirms that increasing layer 2 neurons consistently yields greater gains than increasing layer 1 neurons, suggesting the second TabKAN layer serves as the primary classification discriminator.  Configurations with layer 2 neurons = 64 or layer 2 neurons = 128 yielded uniformly lower performance ($F1 \leq 0.8921$), indicating that layer 2 neuron capacity is the dominant factor governing representational power in this architecture.

\textbf{Effect of Grid Size Configuration.}  \autoref{fig:heatmapgrids} presents the maximum weighted $F1$-score for each combination of layer 1 and layer 2 grid sizes, marginalized over all neuron configurations.  Single-layer grid size sensitivity was moderate, with $F1$ ranging from $0.8400$ (grid = 1) to $0.8649$ (grid = 4), and no monotonic trend with increasing grid size, reflecting saturation of Fourier frequency components beyond a modest grid.  In contrast, two-layer configurations exhibited substantially higher and more consistent performance across grid size combinations, with the majority of cells exceeding $F1 = 0.87$.  The best ($F1 = 0.9038$) was achieved at layer 1 grid size = 6, layer 2 grid size = 7, confirming the selected architecture.  The second-best result (layer 1 grid size = 9, layer 2 grid size = 9; $F1 = 0.8970$) suggests that moderate-to-high grid sizes in both layers generally favor performance, while very low grid sizes in layer 2 (grid size = 1–2) produce weaker results regardless of layer 1 grid size.  This pattern indicates that sufficient Fourier frequency resolution in layer 2 is necessary to capture the nonlinear concept interaction boundaries learned by the Polynomial Feature Expansion module.

Taken together, the sensitivity analysis supports the selected configuration, two TabKAN  layers with 64 neurons and grid size 6 in the first layer, and 512 neurons and grid size 7 in the second layer, as a strong-performing architecture within the explored search space. This configuration achieved the peak weighted $F1$-score of $0.9038$ while maintaining tractable model complexity.

\section{Training and Optimization Details}
\label{appendix:optimization_details}
Optimization was performed using AdamW with decoupled weight decay of 0.1 and an initial learning rate $\eta_0=5\times10^{-3}$.  A manual step-wise decay schedule was applied:
\begin{equation}
\eta(\mathrm{e})=\eta_0 \times 0.1^{1[\mathrm{e}\ge10]+1[\mathrm{e}\ge20]+1[\mathrm{e}\ge40]}
\label{eq:optim}
\end{equation}
yielding $\eta=5\times10^{-3}$ for epochs 0–9, $\eta=5\times10^{-4}$ for epochs 10–19, $\eta=5\times10^{-5}$ for epochs 20–39, and $\eta=5\times10^{-6}$ for epochs $\ge40$.  Training proceeded for a maximum of 100 epochs with early stopping triggered when the average weighted $F1$-score on the validation fold did not improve for 10 consecutive epochs (patience = 10).  Reproducibility was ensured by fixing random seeds across all components: PyTorch CPU (torch.manual\_seed(0)), PyTorch GPU/CUDA (torch.cuda.manual\_seed(0), torch.cuda.manual\_seed\_all(0)), and NumPy (np.random.seed(0)).

\section{Feature Extraction Detail}
\label{appendix:feature_extraction_detail}
The ten concept groups are motivated as physiologically distinct analytic units. Power spectral and entropy features (PSD, MSE) characterize oscillatory and complexity properties of sleep EEG that are known to be sensitive to neurodegenerative changes. Hjorth parameters provide compact time-domain descriptors of signal variance, mobility, and complexity. Sleep spindles and slow oscillations (SOs) reflect thalamocortical and corticohippocampal activity implicated in memory consolidation. LZW compression index captures algorithmic signal complexity. Temporal statistics characterize the distributional shape of the raw EEG waveform. SVD-based features encode cross-channel structure. Demographic covariates (age, education, ethnicity) are included as a structured concept group to allow the model to jointly represent biological and clinical information within the concept bottleneck. The full pipeline yields 1,379 features per subject. The following feature categories were computed from each of the eight EEG channels: two central electrodes (C3, C4), two mastoid electrodes (M1, M2), and four derived channels (C3-M2, C4-M1, C3-LM, C4-LM):
\begin{itemize} 
\item \textbf{Temporal statistics}: signal minimum, maximum, mean, median, root mean square, skewness, kurtosis, and additional waveform characterization metrics (176 features per channel).
\item \textbf{Hjorth parameters} \cite{hjorth1970eeg}: activity (signal variance), mobility (ratio of the standard deviation of the first derivative to signal standard deviation), and complexity (ratio of the mobility of the first derivative to signal mobility) (24 features per channel). 
\item \textbf{Power spectral density (PSD)}: estimated using Welch's method \cite{welch2003use} over $0.3-35 Hz$ with 4-second segments, $50\%$ overlap, and a $50\%$ Tukey window to minimize spectral leakage \cite{harris2005use}.  Both absolute and relative band powers were computed (160 features per channel).
\item \textbf{Lempel-Ziv-Welch (LZW) compression index} \cite{ziv2003compression}: computed on symbolized, coarse-grained signals to quantify algorithmic complexity (8 features per channel).
\item \textbf{Multi-scale entropy (MSE)} \cite{costa2002multiscale, costa2005multiscale}: assessed signal regularity across 20 temporal scales ($1-20 s$) using sample entropy with pattern length $m=2$ and similarity threshold $r=0.15\sigma$ (160 features per channel).
\item \textbf{Sleep spindles} \cite{Purcell2017}: detected via Morlet wavelet decomposition ($11-15 Hz$) with an adaptive threshold of $1.5$ times the baseline spindle power.  Features capture spindle density, total count, mean amplitude, mean duration, mean chirp metric, and mean symmetry metric (640 features per channel).
\item \textbf{Slow oscillations (SOs)} \cite{Purcell2017}: identified via heuristic amplitude thresholding below $1 Hz$.  Features include SO rate, median amplitude, median peak-to-peak amplitude, median duration, and median slope, among others.  Temporal coupling between spindles and SOs was quantified via spindle–SO phase relationships (112 features per channel).
\item \textbf{Singular value decomposition (SVD)}: applied across multi-channel data to yield singular values, explained variance, and cumulative variance metrics (32 features, computed globally across channels).
\item \textbf{SVD component weights}: channel-level contribution weights to each principal component (64 features, computed globally).
\item \textbf{Demographics}: age, years of education, and self-reported ethnicity (3 features, subject-level covariates).
\end{itemize}

\section{Preprocessing Pipeline}
\label{appendix:preprocessing}
Preprocessing followed National Sleep Research Resource harmonization guidelines and was carried out using Luna v0.99, an open-source C/C++ package for large-scale sleep signal analysis.  The pipeline comprised five sequential stages \cite{kozhemiako2022sources}: (1) epoch trimming to isolate sleep periods; (2) exclusion of 30-second epochs coinciding with manually annotated arousal or respiratory events; (3) electrode nomenclature standardization; (4) multi-channel signal generation with consistent resampling and microvolt scaling; and (5) robust normalization using median-centered scaling, where the effective standard deviation was estimated as $\sigma \approx 0.7413 \times IQR$ to mitigate DC offset effects.

Stringent artifact rejection criteria were applied to exclude epochs exhibiting any of the following: flat signal duration exceeding $10\%$ of the epoch; signal clipping exceeding $10\%$ of the epoch; high-amplitude activity exceeding $100 \mu V$ in more than $20\%$ of the epoch or exceeding $250 \mu V$ in more than $1\%$ of the epoch; maximum signal amplitude below $5 \mu V$; or amplitude deviating beyond $\pm3\sigma$ from the normalized epoch mean.  Epoch-level normalization was then applied to each channel by subtracting its median value, following \cite{kozhemiako2022sources}.

\section{Computational Environment and More Baseline Model Details}
\label{appendix:computational_environment}
All experiments were conducted on a server equipped with 40 CPU cores, 480 GB RAM, and an NVIDIA H100 GPU with 160 GB GPU memory.  The software environment comprised Python 3.9.21, PyTorch 2.3.1, and scikit-learn 1.4.2.  The TabKAN and FourierKAN components were implemented natively in PyTorch without reliance on external KAN libraries.

The following baselines were implemented:
\begin{itemize} 
\item \textbf{SVM}: following the approaches of Geng et al. \cite{geng2022sleep} and Aljalal et al. \cite{Aljalal2024}, with kernel function and regularization as tunable hyperparameters.
\item \textbf{KNN}: following Aljalal et al. \cite{diagnostics14151619}, with number of neighbors as the primary hyperparameter.
\item \textbf{Random Forest}: following Rutkowski et al. \cite{rutkowski2024mild}, with number of estimators and maximum depth tuned.
\item \textbf{Bi-LSTM and Bi-GRU}: inspired by Geng et al. \cite{geng2022sleep} and Said and Göker \cite{Said2024}.
\item \textbf{DCNN-SBiL}: inspired by Nirmala Devi et al. \cite{devi2025dcnn}.
\item \textbf{Additional ensemble and linear baselines}: ExtraTrees, XGBoost, LightGBM, GradientBoosting, and Logistic Regression.
\end{itemize}

\section{Robustness Under Class Balancing}
\label{appendix:smote}
\textit{Does the performance advantage of CPTabKAN persist when class imbalance is addressed through synthetic oversampling?}

Under the SMOTE condition, CPTabKAN-Second Order achieved $F1=0.8902 \pm 0.046$ and CPTabKAN-First Order achieved $F1=0.8735 \pm 0.024$, both remaining above the strongest SMOTE baseline (VotingEnsemble: $0.8466 \pm 0.104$; GradientBoosting: $0.8464 \pm 0.092$). Notably, several baselines exhibited pronounced performance degradation under SMOTE: Bi-GRU dropped from 0.8452 to $0.6984 \pm 0.210$, and KNN from 0.8351 to $0.6747 \pm 0.075$. This degradation is consistent with the known sensitivity of recurrent models and distance-based classifiers to synthetic minority interpolations: recurrent architectures may overfit to synthetically generated sequences that do not reflect the true data manifold, while KNN's decision boundaries are directly distorted by the addition of interpolated neighbors. In contrast, CPTabKAN exhibited lower performance variance under both conditions and did not exhibit marked degradation under SMOTE augmentation. These results provide secondary robustness support for the proposed framework's stability under class-balancing interventions, reinforcing the primary findings reported under the no-SMOTE condition.

\section{Cross-Validation and Class Balancing}
\label{appendix:CV-ClassBalancing}
Model performance was evaluated using 10-fold cross-validation with shuffling enabled (n\_splits=10, shuffle=True, random\_state=42). Reproducible data partitioning was ensured through the fixed random state. No held-out test set was used; all performance estimates are derived from cross-validation. Two experimental conditions were assessed: (1) without class balancing, where models were trained on the original imbalanced training fold; and (2) with class balancing, where SMOTE \cite{chawla2002smote1} was applied exclusively within each training fold after splitting, using the default implementation in \texttt{imblearn.over\_sampling.SMOTE} (k\_neighbors=5, random\_state=42), with the oversampling ratio set to achieve full 1:1 class balancing. This prevents synthetic samples from appearing in validation folds, thereby avoiding information leakage. SMOTE generates new MCI samples by linear interpolation between randomly selected minority-class neighbors in the feature space, increasing the effective MCI training samples to match the majority class size.

\section{Ablation Study Results}
\label{appendix:ablation}
Table~\ref{tbl:ablation} summarizes six ablation configurations evaluated under the primary (no-SMOTE) condition. Results are organized by the contribution of each module.

\begin{table}[width=.9\linewidth,cols=5,pos=h]
\caption{Ablation study results for CPTabKAN under the no-SMOTE experimental condition.  CE = Concept Encoders (Module 1); PFE = Polynomial Feature Expansion (Module 2); TabKAN = Tabular Kolmogorov-Arnold Network classifier (Module 3).  \ding{52} = module present; \ding{56} = module absent.  When CE is absent, PFE operates on raw input features.  When TabKAN is absent, an alternative classifier (MLP, Bi-GRU, Bi-LSTM, CNN, dual-branch CNN, or KAN) is substituted; the reported $F1$ is the best result across these alternatives.  Bold values indicate the best performance within each condition, and underlined values indicate the second-best performance.}\label{tbl:ablation}
\begin{tabular*}{\tblwidth}{@{} CCCLL@{} }
\toprule
CE (M1) & PFE (M2) & TabKAN (M3) & Configuration & Weighted F1 \\
\midrule
\ding{52} & \ding{52} & \ding{52} & CPTabKAN-Second Order (proposed) & \textbf{0.9038} \\
\ding{52} & \ding{56} & \ding{52} & CPTabKAN-First Order (proposed) & \underline{{0.9007}} \\
\ding{52} & \ding{52} & \ding{56} & CE + PFE + alt. classifier & 0.8748 \\
\ding{52} & \ding{56} & \ding{56} & CE + alt. classifier & 0.8614 \\
\ding{56} & \ding{52} & \ding{52} & PFE (on raw features) + TabKAN & 0.8496 \\
\ding{56} & \ding{56} & \ding{52} & TabKAN only & 0.8489 \\
\bottomrule
\end{tabular*}
\end{table}

\textbf{Contribution of concept encoders.} Removing CE while retaining PFE and TabKAN, applying polynomial expansion directly to the raw 1,379-dimensional feature vector, yielded $F1 = 0.8496$, a degradation of 5.42 percentage points relative to the full model ($F1 = 0.9038$). Removing both CE and PFE, leaving TabKAN alone on raw features, yielded a nearly identical $F1 = 0.8489$. The near-equivalence of these two configurations indicates that polynomial expansion over unstructured raw features provides no meaningful benefit; without domain-structured compression into the concept bottleneck, the expansion degenerates into an intractable input space rather than an inspectable interaction surface. Concept-level abstraction is therefore the primary load-bearing element of the framework.

\textbf{Contribution of polynomial interaction expansion.} Removing PFE while retaining CE and TabKAN (i.e., CPTabKAN-First Order) resulted in $F1 = 0.9007$, a marginal decrease of 0.0031 from the full model. This incremental but consistent gap confirms that second-order polynomial interactions add discriminative value beyond first-order concept scores, as also reported in Section~\ref{sec:interactions}.

\textbf{Contribution of TabKAN classifier.} Replacing TabKAN with the best-performing alternative classifier while retaining CE and PFE yielded $F1 = 0.8748$, a degradation of 2.90 percentage points. Retaining CE alone with the best alternative classifier yielded $F1 = 0.8614$, a further 1.34-point reduction. This pattern indicates that flexible nonlinear learning over the concept-structured and interaction-expanded representation is a material contributor to performance, beyond what simpler classifiers can achieve on the same input.

Taken together, the ablation evidence indicates that all three modules contribute positively and that their contributions are not reducible to a single generic capacity effect. The CE provides the largest independent contribution by transforming a high-dimensional, heterogeneous feature space into a compact and domain-structured representation; TabKAN adds substantial nonlinear discriminative power over that representation; and PFE provides an additional incremental benefit by making cross-concept interactions explicit. This pattern supports the design hypothesis that concept structuring, interaction modeling, and flexible nonlinear classification are complementary rather than redundant components.




\bibliographystyle{link-elsarticle-num}

\bibliography{cas-refs}

@ARTICLE{geng2022sleep,
    AUTHOR={Geng, Duyan  and Wang, Chao  and Fu, Zhigang  and Zhang, Yi  and Yang, Kai  and An, Hongxia },
    TITLE={Sleep EEG-Based Approach to Detect Mild Cognitive Impairment},
    JOURNAL={Frontiers in Aging Neuroscience},
    VOLUME={Volume 14 - 2022},
    YEAR={2022},
    DOI={10.3389/fnagi.2022.865558},
    ISSN={1663-4365},
}

@ARTICLE{Said2024,
    author={Said, Afrah
    and G{\"o}ker, Hanife},
    title={Spectral analysis and Bi-LSTM deep network-based approach in detection of mild cognitive impairment from electroencephalography signals},
    journal={Cognitive Neurodynamics},
    year={2024},
    month={Apr},
    day={01},
    volume={18},
    number={2},
    pages={597-614},
    issn={1871-4099},
    doi={10.1007/s11571-023-10010-y},
}

@ARTICLE{Aljalal2024,
    author={Aljalal, Majid
    and Aldosari, Saeed A.
    and Molinas, Marta
    and Alturki, Fahd A.},
    title={Selecting EEG channels and features using multi-objective optimization for accurate MCI detection: validation using leave-one-subject-out strategy},
    journal={Scientific Reports},
    year={2024},
    month={May},
    day={30},
    volume={14},
    number={1},
    pages={12483},
    issn={2045-2322},
    doi={10.1038/s41598-024-63180-y},
}

@ARTICLE{diagnostics14151619,
    AUTHOR = {Aljalal, Majid and Aldosari, Saeed A. and AlSharabi, Khalil and Alturki, Fahd A.},
    TITLE = {EEG-Based Detection of Mild Cognitive Impairment Using DWT-Based Features and Optimization Methods},
    JOURNAL = {Diagnostics},
    VOLUME = {14},
    YEAR = {2024},
    NUMBER = {15},
    ARTICLE-NUMBER = {1619},
    PubMedID = {39125495},
    ISSN = {2075-4418},
    DOI = {10.3390/diagnostics14151619}
}

@ARTICLE{rutkowski2024mild,
  
    AUTHOR={Rutkowski, Tomasz M.  and Komendziński, Tomasz  and Otake-Matsuura, Mihoko },
             
    TITLE={Mild cognitive impairment prediction and cognitive score regression in the elderly using EEG topological data analysis and machine learning with awareness assessed in affective reminiscent paradigm},
            
    JOURNAL={Frontiers in Aging Neuroscience},
            
    VOLUME={Volume 15 - 2023},
    
    YEAR={2024},
    
    DOI={10.3389/fnagi.2023.1294139},
    
    ISSN={1663-4365}
}

@ARTICLE{devi2025dcnn,
    title = {{DCNN-SBiL}: EEG signal based mild cognitive impairment classification using compact convolutional network},
    journal = {Expert Systems with Applications},
    volume = {273},
    pages = {126553},
    year = {2025},
    issn = {0957-4174},
    doi = {10.1016/j.eswa.2025.126553},
    author = {A. {Nirmala Devi} and M. Latha}
}

@article{spira2008sleep,
    author = {Spira, Adam P. and Blackwell, Terri and Stone, Katie L. and Redline, Susan and Cauley, Jane A. and Ancoli-Israel, Sonia and Yaffe, Kristine},
    title = {Sleep-Disordered Breathing and Cognition in Older Women},
    journal = {Journal of the American Geriatrics Society},
    volume = {56},
    number = {1},
    pages = {45-50},
    doi = {10.1111/j.1532-5415.2007.01506.x},
    year = {2008}
}

@article{zhang2018national,
    author = {Zhang, Guo-Qiang and Cui, Licong and Mueller, Remo and Tao, Shiqiang and Kim, Matthew and Rueschman, Michael and Mariani, Sara and Mobley, Daniel and Redline, Susan},
    title = {The National Sleep Research Resource: towards a sleep data commons},
    journal = {Journal of the American Medical Informatics Association},
    volume = {25},
    number = {10},
    pages = {1351-1358},
    year = {2018},
    month = {05},
    issn = {1527-974X},
    doi = {10.1093/jamia/ocy064},
}

@misc{nsrr,
  author       = {{NSRR Homepage}},
  howpublished = {\url{https://sleepdata.org/}},
  note = {[accessed 9 February 2023]}
}

@misc{sof,
  author       = {{SOF Online website }},
  howpublished = {\url{https://sofonline.ucsf.edu/}},
  note = {[accessed 16 August 2023]}
}

@article{Purcell2017,
    author={Purcell, S. M.
    and Manoach, D. S.
    and Demanuele, C.
    and Cade, B. E.
    and Mariani, S.
    and Cox, R.
    and Panagiotaropoulou, G.
    and Saxena, R.
    and Pan, J. Q.
    and Smoller, J. W.
    and Redline, S.
    and Stickgold, R.},
    title={Characterizing sleep spindles in 11,630 individuals from the National Sleep Research Resource},
    journal={Nature Communications},
    year={2017},
    month={Jun},
    day={26},
    volume={8},
    number={1},
    pages={15930},
    issn={2041-1723},
    doi={10.1038/ncomms15930},
}

@article {kozhemiako2022sources,
	author = {Kozhemiako, Nataliia and Mylonas, Dimitris and Pan, Jen Q. and Prerau, Michael J. and Redline, Susan and Purcell, Shaun M.},
	title = {Sources of Variation in the Spectral Slope of the Sleep EEG},
	volume = {9},
	number = {5},
	elocation-id = {ENEURO.0094-22.2022},
	year = {2022},
	doi = {10.1523/ENEURO.0094-22.2022},
	publisher = {Society for Neuroscience},
	journal = {eNeuro}
}

@article{Salmi2024,
    author={Salmi, Mabrouka
    and Atif, Dalia
    and Oliva, Diego
    and Abraham, Ajith
    and Ventura, Sebastian},
    title={Handling imbalanced medical datasets: review of a decade of research},
    journal={Artificial Intelligence Review},
    year={2024},
    month={Sep},
    day={02},
    volume={57},
    number={10},
    pages={273},
    issn={1573-7462},
    doi={10.1007/s10462-024-10884-2}
}

@article{app14219863,
    AUTHOR = {Hinojosa Lee, Maria Cristina and Braet, Johan and Springael, Johan},
    TITLE = {Performance Metrics for Multilabel Emotion Classification: Comparing Micro, Macro, and Weighted F1-Scores},
    JOURNAL = {Applied Sciences},
    VOLUME = {14},
    YEAR = {2024},
    NUMBER = {21},
    ARTICLE-NUMBER = {9863},
    ISSN = {2076-3417},
    DOI = {10.3390/app14219863}
}

@article{Hancock2023,
    author={Hancock, John T.
    and Khoshgoftaar, Taghi M.
    and Johnson, Justin M.},
    title={Evaluating classifier performance with highly imbalanced Big Data},
    journal={Journal of Big Data},
    year={2023},
    month={Apr},
    day={11},
    volume={10},
    number={1},
    pages={42},
    issn={2196-1115},
    doi={10.1186/s40537-023-00724-5}
}

@article{chawla2002smote1,
  author       = {Chawla, Nitesh V and Bowyer, Kevin W and Hall, Lawrence O and Kegelmeyer, W Philip},
  title        = {SMOTE: synthetic minority over-sampling technique},
  journal      = {Journal of artificial intelligence research},
  volume       = {16},
  pages        = {321--357},
  year         = {2002},
  doi          = {10.1613/jair.953}
}

@article{hjorth1970eeg,
    title = {EEG analysis based on time domain properties},
    journal = {Electroencephalography and Clinical Neurophysiology},
    volume = {29},
    number = {3},
    pages = {306-310},
    year = {1970},
    issn = {0013-4694},
    doi = {10.1016/0013-4694(70)90143-4},
    author = {Bo Hjorth}
}

@article{welch2003use,
  author={Welch, P.},
  journal={IEEE Transactions on Audio and Electroacoustics}, 
  title={The use of fast Fourier transform for the estimation of power spectra: A method based on time averaging over short, modified periodograms}, 
  year={1967},
  volume={15},
  number={2},
  pages={70-73},
  doi={10.1109/TAU.1967.1161901}
}

@article{harris2005use,
  author={Harris, F.J.},
  journal={Proceedings of the IEEE}, 
  title={On the use of windows for harmonic analysis with the discrete Fourier transform}, 
  year={1978},
  volume={66},
  number={1},
  pages={51-83},
  doi={10.1109/PROC.1978.10837}
}

@article{ziv2003compression,
  author={Ziv, J. and Lempel, A.},
  journal={IEEE Transactions on Information Theory}, 
  title={Compression of individual sequences via variable-rate coding}, 
  year={1978},
  volume={24},
  number={5},
  pages={530-536},
  doi={10.1109/TIT.1978.1055934}
}

@article{costa2002multiscale,
  title = {Multiscale Entropy Analysis of Complex Physiologic Time Series},
  author = {Costa, Madalena and Goldberger, Ary L. and Peng, C.-K.},
  journal = {Phys. Rev. Lett.},
  volume = {89},
  issue = {6},
  pages = {068102},
  numpages = {4},
  year = {2002},
  month = {Jul},
  publisher = {American Physical Society},
  doi = {10.1103/PhysRevLett.89.068102},
}

@article{costa2005multiscale,
  title = {Multiscale entropy analysis of biological signals},
  author = {Costa, Madalena and Goldberger, Ary L. and Peng, C.-K.},
  journal = {Phys. Rev. E},
  volume = {71},
  issue = {2},
  pages = {021906},
  numpages = {18},
  year = {2005},
  month = {Feb},
  publisher = {American Physical Society},
  doi = {10.1103/PhysRevE.71.021906},
}

@article{eslamian2025tabkan,
  title={Tabkan: advancing tabular data analysis using Kolmogorov-Arnold network},
  author={Eslamian, Ali and Afzal Aghaei, Alireza and Cheng, Qiang},
  journal={Machine Learning for Computational Science and Engineering},
  volume={1},
  number={2},
  pages={40},
  year={2025},
  publisher={Springer},
  doi = {10.1007/s44379-025-00042-y},
}

@inproceedings{koh2020concept,
  title = 	 {Concept Bottleneck Models},
  author =       {Koh, Pang Wei and Nguyen, Thao and Tang, Yew Siang and Mussmann, Stephen and Pierson, Emma and Kim, Been and Liang, Percy},
  booktitle = 	 {Proceedings of the 37th International Conference on Machine Learning},
  pages = 	 {5338--5348},
  year = 	 {2020},
  editor = 	 {III, Hal Daumé and Singh, Aarti},
  volume = 	 {119},
  series = 	 {Proceedings of Machine Learning Research},
  month = 	 {13--18 Jul},
  publisher =    {PMLR},
  url = 	 {https://proceedings.mlr.press/v119/koh20a.html}
}

@inproceedings{xu2025enhancing,
    author = {Xu, Jinfeng and Chen, Zheyu and Li, Jinze and Yang, Shuo and Wang, Wei and Hu, Xiping and Ngai, Edith},
    title = {Enhancing Graph Collaborative Filtering with FourierKAN Feature Transformation},
    year = {2025},
    isbn = {9798400720406},
    publisher = {Association for Computing Machinery},
    address = {New York, NY, USA},
    doi = {10.1145/3746252.3760909},
    booktitle = {Proceedings of the 34th ACM International Conference on Information and Knowledge Management},
    pages = {5376–5380},
    numpages = {5},
    location = {Seoul, Republic of Korea},
    series = {CIKM '25}
}

@inproceedings{liu2025kan,
 author = {Liu, Ziming and Wang, Yixuan and Vaidya, Sachin and Ruehle, Fabian and Halverson, James and Soljacic, Marin and Hou, Thomas and Tegmark, Max },
 booktitle = {International Conference on Learning Representations},
 editor = {Y. Yue and A. Garg and N. Peng and F. Sha and R. Yu},
 pages = {70367--70413},
 title = {KAN: Kolmogorov\textendash Arnold Networks},
 doi = {10.48550/arXiv.2404.19756},
 volume = {2025},
 year = {2025}
}

@article{wei2018differential,
    doi = {10.1371/journal.pcbi.1006322},
    author = {Wei, Yina AND Krishnan, Giri P. AND Komarov, Maxim AND Bazhenov, Maxim},
    journal = {PLOS Computational Biology},
    publisher = {Public Library of Science},
    title = {Differential roles of sleep spindles and sleep slow oscillations in memory consolidation},
    year = {2018},
    month = {07},
    volume = {14},
    pages = {1-32},
    number = {7},

}

@article {ng2025bayesian,
article_type = {journal},
title = {Bayesian meta-analysis reveals the mechanistic role of slow oscillation-spindle coupling in sleep-dependent memory consolidation},
author = {Ng, Thea and Noh, Eunsol and Spencer, Rebecca MC},
editor = {Peyrache, Adrien and Marquand, Andre F},
volume = 13,
year = 2025,
month = {oct},
pub_date = {2025-10-08},
pages = {RP101992},
citation = {eLife 2025;13:RP101992},
doi = {10.7554/eLife.101992},
journal = {eLife},
issn = {2050-084X},
publisher = {eLife Sciences Publications, Ltd},
}

@article{Ladenbauer2017zu,
  title    = "Promoting Sleep Oscillations and Their Functional Coupling by
              Transcranial Stimulation Enhances Memory Consolidation in Mild
              Cognitive Impairment",
  author   = "Ladenbauer, Julia and Ladenbauer, Josef and K{\"u}lzow, Nadine
              and de Boor, Rebecca and Avramova, Elena and Grittner, Ulrike and
              Fl{\"o}el, Agnes",
  journal  = "J Neurosci",
  volume   =  37,
  number   =  30,
  pages    = "7111--7124",
  month    =  jun,
  year     =  2017,
  address  = "United States",
  language = "en",
  doi = "10.1523/JNEUROSCI.0260-17.2017"
}

@article{li2024data,
    author = {Li, Wentao and Varatharajah, Yogatheesan and Dicks, Ellen and Barnard, Leland and Brinkmann, Benjamin H and Crepeau, Daniel and Worrell, Gregory and Fan, Winnie and Kremers, Walter and Boeve, Bradley and Botha, Hugo and Gogineni, Venkatsampath and Jones, David T},
    title = {Data-driven retrieval of population-level EEG features and their role in neurodegenerative diseases},
    journal = {Brain Communications},
    volume = {6},
    number = {4},
    pages = {fcae227},
    year = {2024},
    month = {08},
    issn = {2632-1297},
    doi = {10.1093/braincomms/fcae227},
}

@article{campos2026differences,
    
AUTHOR={Campos-Beltrán, Diana  and Zhang, Shu  and Marshall, Lisa },
           
TITLE={Differences in sleep spindles and polysomnography in humans: a meta-analysis on the influence of age, sex, and cognitive ability},
          
JOURNAL={Frontiers in Sleep},
          
VOLUME={Volume 5 - 2026},
  
YEAR={2026},
  
DOI={10.3389/frsle.2026.1802882},
  
ISSN={2813-2890}
}

@article{niethard2023aging,
    author = {Niethard, Niels},
    title = {Aging impairs the temporal clustering of sleep spindles},
    journal = {Sleep},
    volume = {46},
    number = {5},
    pages = {zsad011},
    year = {2023},
    month = {05},
    issn = {0161-8105},
    doi = {10.1093/sleep/zsad011},
}

@article {zust2023hierarchy,
	author = {Z{\"u}st, Marc Alain and Mikutta, Christian and Omlin, Ximena and DeStefani, Tatjana and Wunderlin, Marina and Zeller, C{\'e}line Jacqueline and Feh{\'e}r, Kristoffer Daniel and Hertenstein, Elisabeth and Schneider, Carlotta L. and Teunissen, Charlotte Elisabeth and Tarokh, Leila and Kl{\"o}ppel, Stefan and Feige, Bernd and Riemann, Dieter and Nissen, Christoph},
	title = {The Hierarchy of Coupled Sleep Oscillations Reverses with Aging in Humans},
	volume = {43},
	number = {36},
	pages = {6268--6279},
	year = {2023},
	doi = {10.1523/JNEUROSCI.0586-23.2023},
	publisher = {Society for Neuroscience},
	issn = {0270-6474},
	journal = {Journal of Neuroscience}
}

\end{document}